\pdfoutput=1
\documentclass[11pt]{article}

\usepackage{graphicx}
\usepackage{subcaption}

\usepackage{EACL2023}
\usepackage{multirow}
\usepackage{subcaption}

\usepackage{times}
\usepackage{latexsym}
\usepackage{booktabs}
\usepackage{amsmath}% For proper rendering and hyphenation of words containing Latin characters (including in bib files)
\usepackage[T1]{fontenc}
\usepackage[utf8]{inputenc}

\usepackage{microtype}
\usepackage{graphicx}

\usepackage{inconsolata}

\newcommand{\jjtext}[1]{\textcolor{black}{#1}}

\title{Revisiting the Markov Property for Machine Translation}

\author{Cunxiao Du \\ Singapore Management University\\ 80 Stamford Rd, Singapore 178902 \\ \texttt{cnsdunm@gmail.com} \And
Hao Zhou\\ Insititute for AI Industry Research (AIR) \\ Tsinghua University \\ \texttt{haozhou0806@gmail.com}
         \AND
        Zhaopeng Tu\\ Tencent AI Lab \\  \texttt{tuzhaopeng@gmail.com} \And
        Jing Jiang \\ Singapore Management University  \\ \texttt{jingjiang@smu.edu.sg}}

\begin{document}
\maketitle
\begin{abstract}
In this paper, we re-examine the Markov property in the context of neural machine translation. 
We design a Markov Autoregressive Transformer~(MAT) and undertake a comprehensive assessment of its performance across four WMT benchmarks. 
Our findings indicate that MAT with an order larger than 4 can generate translations with quality on par with that of conventional autoregressive transformers. 
In addition, counter-intuitively, we also find that the advantages of utilizing a higher-order MAT do not specifically contribute to the translation of longer sentences.
\end{abstract}

\section{Introduction}
\label{sec:intro}

Markov models are classic probabilistic graphical models based on the Markov property.
The Markov property reduces computation complexity and thus makes Markov models highly appealing.
% through the Markov property. 
Markov models have been extensively used in many NLP tasks such as part-of-speech tagging~\cite{ma-hovy-2016-crf, shao-etal-2017-character-crf-pos} and dependency parsing~\cite{Zhang2020secordercrf, Zhang2020FastCRFdependencyparsing}. 
Statistical machine translation~(SMT) has also employed Markov models, e.g., ~\citet{Lavergne2011smtcrf}.

However, with the rise of deep learning in machine translation, autoregressive models~\cite{seq2seq, seq2seqattention, gehring2017convolutional}, particularly autoregressive transformers~\cite{transformer}, have gradually become mainstream. 
During decoding, autoregressive models rely on all the previous tokens.
As a result, they can model long-range dependencies and are thus considered to have superior abilities to express token dependency than Markov models.
The performance of recent advanced Markov models~\cite{wang-etal-2018-neural-hmm, natcrf, Deng2020Cascadedmarkovnat} in MT
are also significantly lower than those of the autoregressive model.

The Markov property dictates that, during decoding, each token can only observe the previous $k$ tokens. 
This characteristic is a considerable drawback for generation tasks that require long contexts, such as story generation. 
However, we believe that in translation, since the source sentence is fully visible, introducing the Markov property on the decoder side might not greatly affect translation performance.

To investigate this hypothesis, we introduce the Markov Autoregressive Transformer (MAT) and evaluate its performance on translation. 
MAT possesses two main features: 1) minimal modifications to autoregressive transformers, and 2) support for high-order Markov models. 
Specifically, the key idea of the $k$th-order Markov property is that the next output token by the model is only dependent on the previous $k$ tokens. 
\jjtext{In this paper, we point out} that this objective can be achieved with a simple modification to the causal mask in the decoder part. 
In contrast to previous Markov models, this simple modification ensures that our MAT has only marginal alterations compared to the autoregressive transformer. 
This allows us to effectively isolate and examine the effects of the Markov property in a manner akin to a controlled variable experiment. 
In addition to the aforementioned benefit, this straightforward modification also enables us to train MAT in parallel, like the vanilla transformer.

We evaluate MAT on several WMT benchmarks and make the following observations:
\begin{itemize}
    \item The first-order Markov property significantly impairs model performance. For instance, on the WMT14 EN-DE task, there is a decline of approximately 3.4 BLEU points~(\S \ref{sec: all}).
    \item For the $k$th-order Markov property, as $k$ increases, the performance of the model becomes increasingly comparable to that of an autoregressive model (e.g., when $k$=5)~(\S \ref{sec: order}).
    \item The benefits of a larger $k$ are not necessarily specific to longer sentences~(\S \ref{sec: length}).
\end{itemize}

In addition to the aforementioned findings, we also discover that MAT also enjoys the following advantages: 1) Linear complexity of attention. To generate a sentence with the length of $n$, the complexity of attention is only $O(kn)$ compared with $O(n^2)$ in vanilla autoregressive transformers. 
For a sample length of 25, the computation for decoder self-attention is reduced by approximately threefold.
2) Key-Value cache free inference. Because MAT only attends to the embeddings of the previous $k$ tokens, it does not require caching any keys and values of the previous tokens during inference. This reduces the memory bandwidth required by the cache at the decoding stage. 
 By limiting the dependence on a fixed number of preceding tokens, the Markov property can potentially simplify the translation model, thereby reducing complexity and computational requirements. 
This might lead to a balance where adequate performance can be achieved more efficiently.

\section{Preliminaries}

\paragraph{Task Definition.} Machine translation aims to translate an input sentence \( X \) in a source language into an output sentence \( Y \) in a target language. The detailed definition is provided in the Appendix~\ref{apd: task}.

\paragraph{Markov Property}
% paragraph / preliminary / section ??
The Markov property~\cite{markov1954theory} is a stochastic property that states that the probability of a future state depends only on the current state and not on the sequence of states that preceded it. 
For MT, mathematically, given {a source sentence $X$ and a sequence of previously generated target tokens $y_1, y_2, \ldots, y_{n-1}$, 
% a sequence of tokens $y_1, y_2, \cdots, y_n$ and the source sequence $X$, 
and the $k$-order Markov properties allow for longer-distance dependencies, as described by the following:
\begin{equation*}
P(y_{n} | X, y_1, y_2, \cdot, y_{n - 1}) = P(y_{n} | X, y_{n-k}, \cdot, y_{n - 1}).
\end{equation*}

% the first-order Markov property can be defined as
% \begin{equation*}
% P(y_n | X, y_1, y_2, \dots, y_{n-1}) = P(y_{n} | X, y_{n-1}).
% \end{equation*}
% More generally, high-order
\section{Markov Autoregressive Transformer (MAT)}

\subsection{Overview}

Our MAT consists of two parts: 1) an Encoder, and 2) a Markov Decoder. 
We keep the Encoder the same as in the vanilla transformer. 
For the Markov Decoder, the only difference lies in the attention mechanism, which is elaborated as follows.

\subsection{Markov Attention Mechanism}
To keep the Markov property in the decoder, we use a mechanism called transparent Markov attention.
To be specific, Markov attention has two characteristics:
\begin{itemize}
    \item $k$-Order Attention Mask. To prevent the current token from accessing the information beyond what the Markov property allows, we may use a lower triangular matrix to only keep the attention weights within the window size $k$. However, it is worth noting that using this kind of mask alone does not guarantee that information will not leak~\cite{ngramtransformer}. This is because as the number of layers $L$ increases, the current token will encompass information from the former tokens than $k$, violating the Markov property of only observing the previous $k$ tokens. A clearer example is provided in the Appendix~\ref{apd: leak}.
    \item Transparent Attention. Inspired by the two-stream attention~\cite{Yang2019XLNetGA}, we propose a simple method called Transparent Attention to fix the information leakage in the $k$-Order Attention Mask. With such attention, the keys and values of previous tokens are not updated, i.e., they are always set to be the static word embeddings of the corresponding tokens.
    %Although this seems to diminish the expressive power of attention, in our experiments, this simple method does not perform much worse. Moreover, under this approach, we can still perform training in parallel during the training process.
\end{itemize}

\begin{figure}[t]
\centering

\begin{minipage}{0.23\textwidth}
\includegraphics[width=\linewidth]{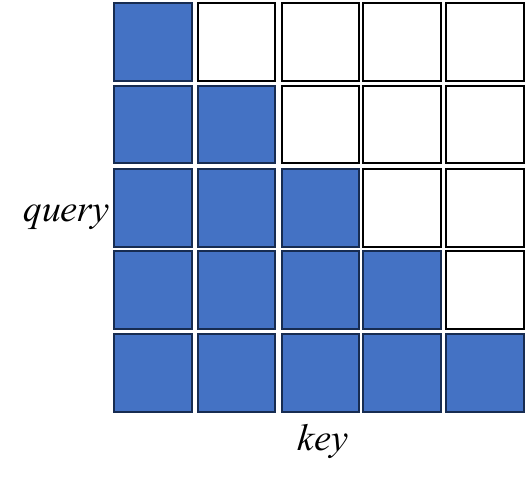}
\end{minipage}
\begin{minipage}{0.23\textwidth}
\includegraphics[width=\linewidth]{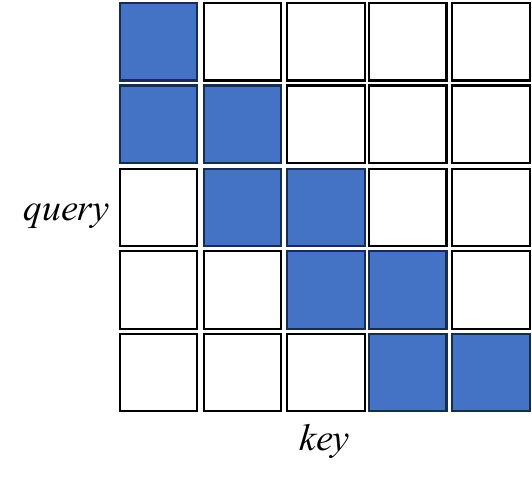}
\end{minipage}

\caption{The illustration of the original casual attention mask (left) and $second$-Order Attention Mask (right).}
\end{figure}

% \subsection{KV-Cache Free Inference}
% Our MAT can also be viewed as a special version of the autoregressive model, so we keep the inference strategy the same as autoregressive translation models, i.e., beam search with length penalty. However, due to the modifications we made to the decoder attention, it is only necessary to attend to the embeddings of the previous tokens. Consequently, there is no need to cache the keys and values of all previous tokens as done in prior methods. This reduces the memory bandwidth required by the transformer at the inference stage by a factor of $O(T)$.
\section{Experiments}
\label{sec:exp}

\subsection{Data}

We conduct experiments on major benchmark MT datasets at different scales that are widely used in previous studies: WMT14 English$\Leftrightarrow$German (En$\Leftrightarrow$De, 4.5M pairs), and large-scale WMT17 English$\Leftrightarrow$Chinese (En$\Leftrightarrow$Zh, 20M pairs). 
For fair comparison, we report BLEU scores~\cite{papineni2002bleu} on En$\Leftrightarrow$De and Zh$\Rightarrow$En, and Sacre BLEU scores~\cite{post2018call} on En$\Rightarrow$Zh. 
  % \jjcomment{What's the reason of using BLEU on some pairs and Sacre BLEU on other pairs? Is this following previous practice?} \cxdu{yes prof, it is a very common setting}
The other details can be found in Appendix~\ref{apd: data}.
 
% We preprocessed the datasets with a joint BPE~\cite{Sennrich:BPE} with 32K merge operations for En$\Leftrightarrow$De, and 32K bpe for En$\Leftrightarrow$Zh.
% For the fair comparison, We reported the BLEU~\cite{papineni2002bleu} on En$\Leftrightarrow$De and Zh$\Rightarrow$En, and Sacre BLEU~\cite{post2018call} on En$\Rightarrow$Zh.
\begin{figure}[htbp]
\centering
\includegraphics[width=\linewidth]{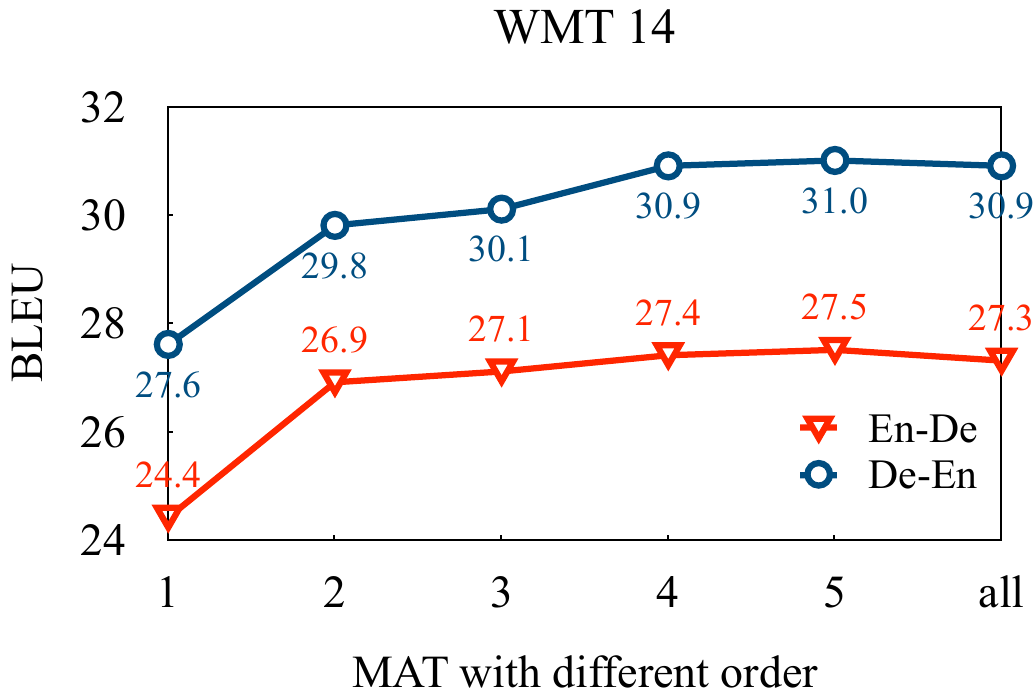}
\caption{In the WMT14 EN-DE dataset, experimental results for MAT with varying values of $k$. It indicates that as $k$ increases, the BLEU score for MAT exhibits an upward trend. However, the improvements plateau when $k$ exceeds 3.}
\label{fig: different_order}
\end{figure}
\begin{table*}[t]
\centering
\begin{tabular}{l ll ll}
\toprule
\multirow{2}{*}{\bf Model} & \multicolumn{2}{c}{\textbf{WMT14}} & \multicolumn{2}{c}{\textbf{WMT17}} \\
 \cmidrule(lr){2-3}\cmidrule(lr){4-5}
 & \textbf{En-De} &\textbf{De-En} &\textbf{En-Zh}  &  \textbf{Zh-En}\\
\midrule
% \bf {Autoregressive} \\
{\bf Autoregressive} Transformer~\cite{transformer}   &  27.8 & 31.3 &  34.4 & 24.0\\
{\bf Autoregressive} Transparent Transformer    &  27.3 & 31.2 &  33.9 & 23.3\\

\midrule

\bf Markov Models \\
Bigram CRF~\cite{natcrf}     & 23.4 & 27.2 & - & -\\
Non-autoregressive Markov Transformer~\cite{Deng2020Cascadedmarkovnat} &24.4 &29.4 & - & - \\
Autoregressive Markov Transformer (Ours, $k$=5) & 27.5 &31.0 & 33.9 & 23.3 \\
\bottomrule
\end{tabular}
\caption{BLEU scores on two benchmarks.
}
\label{tab:main}
\end{table*}

\subsection{Baselines}
To investigate the impact of the Markov property on model performance, we consider the following models as our baselines: 1) Standard Autoregressive Transformer, \jjtext{which attends to \emph{all} previous tokens}, 2) Transparent Attention Transformer, i.e., the transformer with transparent attention, \jjtext{which attends to the contextualized embeddings of the previous $k$ tokens}, and 3) two other Markov Translation Models as reference points. The details of these two models can be found at Appendix~\ref{apd: related}.
\begin{figure*}[htbp]
\centering

\begin{subfigure}{\textwidth}
    \centering
    \includegraphics[width=\textwidth]{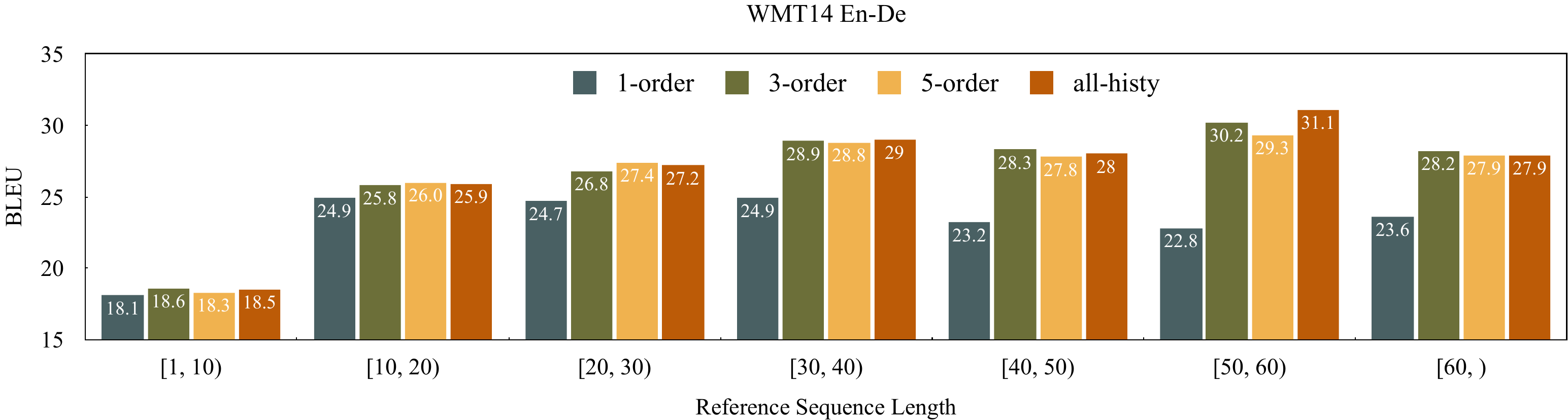}
    \label{fig:sub:different_order}
\end{subfigure}

% \vspace{1cm} % Adds some space between the subfigures

\begin{subfigure}{\textwidth}
    \centering
    \includegraphics[width=\textwidth]{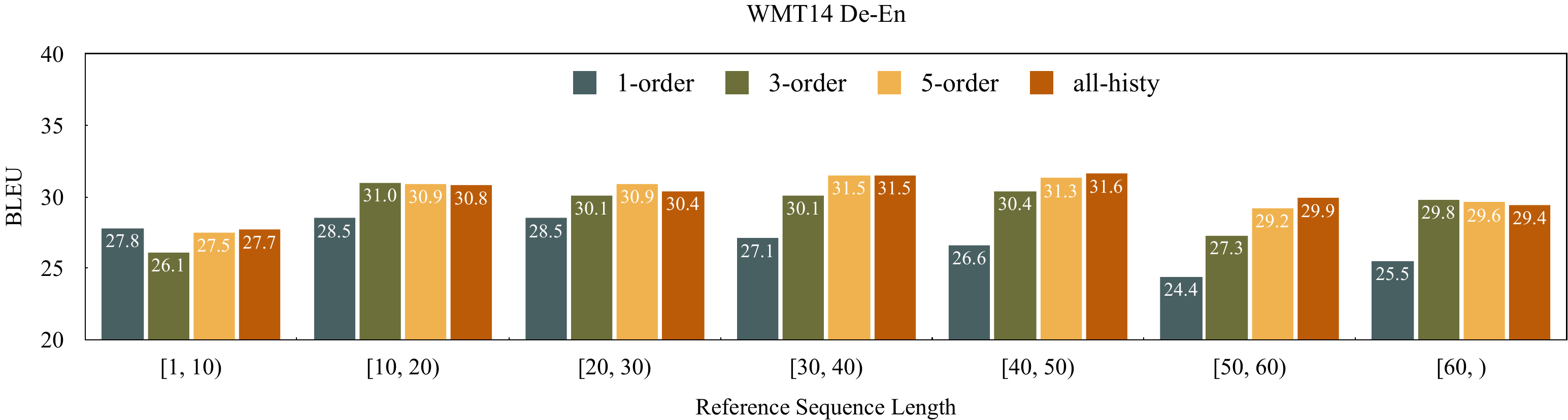}
    \label{fig:sub:different_length}
\end{subfigure}
\vspace{-1cm}

\caption{Performance of the generated translations with respect to
the lengths of the reference sentences.}
\label{fig:sub:different_length}
\vspace{-0.2cm}
\end{figure*}
% \begin{itemize}
%     \item Standard Autoregressive Transformer. The Standard Autoregressive Transformer adopts the Transformer \textsc{Base} architecture, consisting of 6 encoder layers, 6 decoder layers, 8 attention heads, 512 model dimensions, and 2048 hidden dimensions.
%     \item Transparent Attention Transformer. The Transparent Attention Transformer replaces the decoder self-attention in the standard autoregressive Transformer with transparent attention, while all other components remain unchanged. This model provides a platform to rigorously examine the side effects introduced by transparent attention and to precisely investigate the impact of Markov property in MAT.
%     \item Bigram CRF. The Bigram CRF employs the Linear-CRF as its decoder while leveraging the standard Transformer Encoder as the encoder part. More specifically, Bigram CRF utilizes a non-autoregressive Transformer decoder to model $P(y_i|x, pos_{i})$. Subsequently, it deploys a low-rank matrix $M \in |V|^2$ to represent the transition probabilities between adjacent tokens, thereby achieving first-order Markov property. See  for detail.
%     \item  Cascade Markov Transformer. See  for detail.
% \end{itemize}

\subsection{Results}
\label{sec: all}

\jjtext{Comparison between our MAT model and the baselines is shown in Table~\ref{tab:main}.
From the table, we observe the following:}
\begin{itemize}
    \item \textit{Transparent Attention slightly decreases the BLEU score of the model.} \jjtext{Comparing Autoregressive Transformer and Autoregressive Transparent Transformer,} it is evident that employing transparent attention leads to an average performance drop of approximately 0.3 on the WMT14 En$\Leftrightarrow$De benchmark and about 0.6 on the WMT17 En$\Leftrightarrow$Zh benchmark, \jjtext{which is not substantial}.
    
    %This is attributed to the fact that transparent attention does not allow the current token to access the key and value representations of previous tokens, limiting it to attend only to their corresponding word embeddings. Nonetheless, this decrement in performance is relatively modest, even on the large-scale dataset WMT17 En$\Leftrightarrow$Zh. We speculate that this might be due to the encoder effectively capturing sufficient contextual information.
    
    \item \textit{MAT demonstrates significant improvement over previous Markov models.} Compared to previous Markov models \jjtext{for MT, i.e., Bigram CRF and Non-autoregressive Markov Transformer}, we observe that on the  WMT14 En$\Leftrightarrow$De dataset, MAT, with the same \jjtext{model} size, achieves an improvement of 2-3 BLEU points. 
    Notably, the order choice of MAT is 5, consistent with the Non-autoregressive Markov Transformer. This, in fact, suggests that the Markov property is not the primary reason for the 
    % inferior 
    \jjtext{relatively low} performance of earlier Markov models. For the Bigram CRF model, we postulate that one primary limitation is its sole reliance on first-order Markov properties. 
    Furthermore, modeling the relationship between tokens (i.e., the transition matrix) using a low-rank matrix might also contribute to its performance degradation.
    % \jjcomment{Currently there's not a related work section introducing the two earlier Markov models. I'm afraid that here readers many not understand what you are referring to (e.g., low-rank matrix) if they are not familiar with the Bigram CRF model.}
Regarding the Non-autoregressive~\cite{NAT, Du2021OAXE} Markov Transformer, we hypothesize that the main reason for its performance decline might be the pruning during inference through a lower-order Markov model, resulting in the absence of suitable candidates within the candidate set.
    \item \textit{MAT achieves performance comparable to the standard Autoregressive Transformer, albeit slightly worse.} We observe that the performance of MAT slightly decreases compared to the standard Autoregressive Transformer. However, compared with the transparent autoregressive Transformer, MAT's performance remains almost the same. This suggests that within the current MAT architecture, employing the 5-order Markov property does not compromise its translation capabilities.
    
    % In Section ~\ref{sec: order}, we will dive deeper into the impact of different orders on translation quality.
\end{itemize}

\subsection{Analysis}

\paragraph{MAT with Different Order}
\label{sec: order}
% In order to dive deeper into the Markov property, we have chosen to investigate the impact of different orders on MAT within this subsection. 

Recall that in our MAT model with $k$th-order Markov property, $k$ indicates MAT's ability to process previous tokens. 
An intuitive hypothesis is that a larger $k$ might yield better performance \jjtext{because it captures a longer context}. 
However, \jjtext{we find that} empirical results do not fully align with it. 
In Figure~\ref{fig: different_order}, \jjtext{we plot the performance with respect to different values of $k$}. 
We find the following three observations:
1) At $k$=1, the model's performance sees a significant drop compared to a non-Markov model. 
One potential reason is that the complexity of the translation data far exceeds what a first-order Markov model can encapsulate, and another reason is the self-attention in the transformer decoder is no longer useful. 
% In fact, it degenerates into a fully-connected layer. 
Therefore, the decline may also be related to the architecture of the transformer. 
%In the future, there may be deep learning models specifically designed to adapt to the first-order Markov property to better assist in generation.
2) When $k$ is in the range of 2-4, increasing $k$ provides noticeable gains. 
This phenomenon is evident across datasets from both directions.
3) For $k$ values greater than 4, further increasing $k$ does not result in significant performance improvements. % explain why

\paragraph{MAT for References of Different Lengths}

\label{sec: length}
We further examine the impact of different reference lengths on MAT's performance in Figure~\ref{fig:sub:different_length}. 

%Insights from Figure~\ref{fig:sub:different_length} reveals the following:

For $k$=1, there is a noticeable degradation in performance across all sentence lengths. 
This observation \jjtext{is consistent with} previous experiments.

Interestingly, the advantages of a higher-order MAT do not always become more pronounced in longer sentences. 
For instance, in the WMT14 en-de results, the 3rd-order MAT consistently outperforms the 5th-order MAT for sample buckets with sentence lengths over 40. 
This is counter-intuitive because as a sentence gets longer, a higher-order Markov model, with its ability to access a broader previous context, supposedly would be able to utilize more information and give better results.

This unexpected phenomenon might be attributed to particular linguistic characteristics of the target language. 
This theory gains traction when looking at the WMT14 de-en results, where the 3rd-order MAT is only better than the 5th-order MAT in buckets with sentence lengths beyond 60.

\section{Conclusions}

In this paper, we re-examine the Markov property in machine translation. 
We design an experimental Markov model based on the transformer architecture. 
We verify that higher-order Markov properties have a very slight impact on the model's translation quality. 
Moreover, we find that longer sentences do not necessarily require higher-order Markov models. 
In the future, we aim to design faster and more lightweight models to leverage the advantages of the Markov property. And also extend this idea to large language model and other tasks needs real-time decoding like rumor detection~\cite{xuan2023aacl} and infodemic surveillance~\cite{xuanipm2023}.

\section{Limitations}

In this article, we primarily explore the impact of the Markov property on model translation quality. 
\jjtext{We acknowledge that there are still several limitations of our study:}
1) Compared to other Markov models, e.g., bigram CRF, our model cannot generate translations in parallel (i.e., in a non-autoregressive manner). 
Although our model can achieve acceleration compared to the standard autoregressive transformer, we have not fully explored the potential of Markov models in parallel generation. 
2) Our current experiments are based on the transformer, neglecting other architectures, such as CNNs~\cite{wu2019payless} or advanced RNNs~\cite{sun2023retentive}. 
Markov models might perform better on RNN translation models.
3) Regarding the scaling laws~\cite{ghorbani2021scaling} for Markov models, due to our limited GPU resources, we are unable to further explore Markov models of different sizes. 
If more resources become available in the future, it might be meaningful to investigate the performance of scaling laws within Markov models.

\section*{Acknowledgement}
We thank the anonymous reviewers for their helpful comments during the review of this paper. The first author wants to give special thanks to Songlin Yang from MIT, because she encouraged him to transform the idea into a paper. 
This work is partially supported by the Natural Science Foundation of China (62376133).

\bibliography{anthology,custom}

\begin{thebibliography}{26}
\expandafter\ifx\csname natexlab\endcsname\relax\def\natexlab#1{#1}\fi

\bibitem[{Bahdanau et~al.()Bahdanau, Cho, and Bengio}]{seq2seqattention}
Dzmitry Bahdanau, Kyunghyun Cho, and Yoshua Bengio.
\newblock Neural machine translation by jointly learning to align and
  translate.
\newblock In \emph{ICLR}.

\bibitem[{Chelba et~al.(2020)Chelba, Chen, Bapna, and
  Shazeer}]{ngramtransformer}
Ciprian Chelba, Mia Chen, Ankur Bapna, and Noam Shazeer. 2020.
\newblock \href {http://arxiv.org/abs/2001.04451} {Faster transformer decoding:
  N-gram masked self-attention}.

\bibitem[{Deng et~al.(2021)Deng, Awadallah, Meek, Polozov, Sun, and
  Richardson}]{deng-etal-2021-structure}
Xiang Deng, Ahmed~Hassan Awadallah, Christopher Meek, Oleksandr Polozov, Huan
  Sun, and Matthew Richardson. 2021.
\newblock \href {https://doi.org/10.18653/v1/2021.naacl-main.105}
  {Structure-grounded pretraining for text-to-{SQL}}.
\newblock In \emph{Proceedings of the 2021 Conference of the North American
  Chapter of the Association for Computational Linguistics: Human Language
  Technologies}, pages 1337--1350, Online. Association for Computational
  Linguistics.

\bibitem[{Deng and Rush(2020)}]{Deng2020Cascadedmarkovnat}
Yuntian Deng and Alexander~M. Rush. 2020.
\newblock Cascaded text generation with markov transformers.
\newblock In \emph{NeurIPS}.

\bibitem[{Du et~al.(2021)Du, Tu, and Jiang}]{Du2021OAXE}
Cunxiao Du, Zhaopeng Tu, and Jing Jiang. 2021.
\newblock Order-agnostic cross entropy for non-autoregressive machine
  translation.
\newblock In \emph{Proc. of ICML}.

\bibitem[{Gehring et~al.(2017)Gehring, Auli, Grangier, Yarats, and
  Dauphin}]{gehring2017convolutional}
Jonas Gehring, Michael Auli, David Grangier, Denis Yarats, and Yann~N Dauphin.
  2017.
\newblock Convolutional sequence to sequence learning.
\newblock In \emph{ICML}.

\bibitem[{Ghorbani et~al.(2021)Ghorbani, Firat, Freitag, Bapna, Krikun, Garcia,
  Chelba, and Cherry}]{ghorbani2021scaling}
Behrooz Ghorbani, Orhan Firat, Markus Freitag, Ankur Bapna, Maxim Krikun,
  Xavier Garcia, Ciprian Chelba, and Colin Cherry. 2021.
\newblock Scaling laws for neural machine translation.
\newblock In \emph{ICLR}.

\bibitem[{Gu et~al.(2018)Gu, Bradbury, Xiong, Li, and Socher}]{NAT}
Jiatao Gu, James Bradbury, Caiming Xiong, Victor~OK Li, and Richard Socher.
  2018.
\newblock Non-autoregressive neural machine translation.
\newblock In \emph{ICLR}.

\bibitem[{Lavergne et~al.(2011)Lavergne, Allauzen, Crego, and
  Yvon}]{Lavergne2011smtcrf}
Thomas Lavergne, A.~Allauzen, Josep~Maria Crego, and François Yvon. 2011.
\newblock From n-gram-based to crf-based translation models.
\newblock In \emph{WMT@EMNLP}.

\bibitem[{Ma and Hovy(2016)}]{ma-hovy-2016-crf}
Xuezhe Ma and Eduard Hovy. 2016.
\newblock End-to-end sequence labeling via bi-directional {LSTM}-{CNN}s-{CRF}.
\newblock In \emph{ACL}.

\bibitem[{Markov(1954)}]{markov1954theory}
A.~A. Markov. 1954.
\newblock \emph{Theory of Algorithms}.
\newblock Academy of Sciences of the USSR.

\bibitem[{Papineni et~al.(2002)Papineni, Roukos, Ward, and
  Zhu}]{papineni2002bleu}
Kishore Papineni, Salim Roukos, Todd Ward, and Wei-Jing Zhu. 2002.
\newblock Bleu: a method for automatic evaluation of machine translation.
\newblock In \emph{ACL}.

\bibitem[{Post(2018)}]{post2018call}
Matt Post. 2018.
\newblock A call for clarity in reporting bleu scores.
\newblock In \emph{WMT}.

\bibitem[{Sennrich et~al.(2016)Sennrich, Haddow, and Birch}]{Sennrich:BPE}
Rico Sennrich, Barry Haddow, and Alexandra Birch. 2016.
\newblock Neural machine translation of rare words with subword units.
\newblock In \emph{ACL}.

\bibitem[{Shao et~al.(2017)Shao, Hardmeier, Tiedemann, and
  Nivre}]{shao-etal-2017-character-crf-pos}
Yan Shao, Christian Hardmeier, J{\"o}rg Tiedemann, and Joakim Nivre. 2017.
\newblock Character-based joint segmentation and {POS} tagging for {C}hinese
  using bidirectional {RNN}-{CRF}.
\newblock In \emph{IJCNLP}.

\bibitem[{Sun et~al.(2023)Sun, Dong, Huang, Ma, Xia, Xue, Wang, and
  Wei}]{sun2023retentive}
Yutao Sun, Li~Dong, Shaohan Huang, Shuming Ma, Yuqing Xia, Jilong Xue, Jianyong
  Wang, and Furu Wei. 2023.
\newblock Retentive network: A successor to transformer for large language
  models.
\newblock \emph{arXiv preprint arXiv:2307.08621}.

\bibitem[{Sun et~al.(2019)Sun, Li, Wang, Lin, He, and Deng}]{natcrf}
Zhiqing Sun, Zhuohan Li, Haoqing Wang, Zi~Lin, Di~He, and Zhi-Hong Deng. 2019.
\newblock Fast structured decoding for sequence models.
\newblock In \emph{NeurIPS}.

\bibitem[{Sutskever et~al.(2014)Sutskever, Vinyals, and Le}]{seq2seq}
Ilya Sutskever, Oriol Vinyals, and Quoc~V Le. 2014.
\newblock Sequence to sequence learning with neural networks.
\newblock \emph{Advances in neural information processing systems}, 27.

\bibitem[{Vaswani et~al.(2017)Vaswani, Shazeer, Parmar, Uszkoreit, Jones,
  Gomez, Kaiser, and Polosukhin}]{transformer}
Ashish Vaswani, Noam Shazeer, Niki Parmar, Jakob Uszkoreit, Llion Jones,
  Aidan~N Gomez, Lukasz Kaiser, and Illia Polosukhin. 2017.
\newblock Attention is all you need.
\newblock In \emph{NeurIPS}.

\bibitem[{Wang et~al.(2018)Wang, Zhu, Alkhouli, Gan, and
  Ney}]{wang-etal-2018-neural-hmm}
Weiyue Wang, Derui Zhu, Tamer Alkhouli, Zixuan Gan, and Hermann Ney. 2018.
\newblock Neural hidden {M}arkov model for machine translation.
\newblock In \emph{ACL}.

\bibitem[{Wu et~al.(2019)Wu, Fan, Baevski, Dauphin, and Auli}]{wu2019payless}
Felix Wu, Angela Fan, Alexei Baevski, Yann~N Dauphin, and Michael Auli. 2019.
\newblock Pay less attention with lightweight and dynamic convolutions.
\newblock In \emph{ICLR}.

\bibitem[{Yang et~al.(2019)Yang, Dai, Yang, Carbonell, Salakhutdinov, and
  Le}]{Yang2019XLNetGA}
Zhilin Yang, Zihang Dai, Yiming Yang, Jaime~G. Carbonell, Ruslan Salakhutdinov,
  and Quoc~V. Le. 2019.
\newblock Xlnet: Generalized autoregressive pretraining for language
  understanding.
\newblock In \emph{NeurIPS}.

\bibitem[{Zhang and Gao(2023)}]{xuan2023aacl}
Xuan Zhang and Wei Gao. 2023.
\newblock Towards llm-based fact verification on news claims with a
  hierarchical step-by-step prompting method.
\newblock \emph{arXiv preprint arXiv:2310.00305}.

\bibitem[{Zhang and Gao(2024)}]{xuanipm2023}
Xuan Zhang and Wei Gao. 2024.
\newblock Predicting viral rumors and vulnerable users with graph-based neural
  multi-task learning for infodemic surveillance.
\newblock \emph{Information Processing \& Management}.

\bibitem[{Zhang et~al.(2020{\natexlab{a}})Zhang, Li, and
  Zhang}]{Zhang2020secordercrf}
Yu~Zhang, Zhenghua Li, and Min Zhang. 2020{\natexlab{a}}.
\newblock Efficient second-order treecrf for neural dependency parsing.
\newblock In \emph{ACL}.

\bibitem[{Zhang et~al.(2020{\natexlab{b}})Zhang, Zhou, and
  Li}]{Zhang2020FastCRFdependencyparsing}
Yu~Zhang, Houquan Zhou, and Zhenghua Li. 2020{\natexlab{b}}.
\newblock Fast and accurate neural crf constituency parsing.
\newblock \emph{ArXiv}, abs/2008.03736.

\end{thebibliography}

\appendix

% \section{Example Appendix}
% 

\appendix
\section{Appendix}
\subsection{Task Definition}
\label{apd: task}
Given a sentence \( X \) in a source language, machine translation aims to produce a sentence \( Y \) in a target language that has the same semantic meaning as \( X \).
Formally, an MT system attempts to output the best translation $Y^{*}$:
\begin{eqnarray*}
Y^{*} = \text{argmax}_{Y} P_\theta(Y|X), 
\end{eqnarray*}
where \( P_\theta(Y|X) \) is the probability of translation \( Y \) given source \( X \).

Autoregressive neural machine translation~(NMT) decomposes \( P(Y|X) \)  by predicting one token (e.g., a subword) of the target sequence at one time, conditioned on the entire source sequence and all previously predicted tokens in the target sequence. 

Formally, given a source sequence \( X = [x_1, x_2, ..., x_{m}] \) and a target sequence \( Y = [y_1, y_2, ..., y_{n}] \), the model is trained to maximize the conditional probability:
\begin{eqnarray*}
P(Y|X) = \prod_{i=1}^{n} P(y_i|X, y_1, ..., y_{i-1}). 
\end{eqnarray*}

\subsection{Information Leakage in $k$-Order Attention Mask}
\label{apd: leak}
A second-order Markov property requires that only the two previous tokens, i.e., \textbf{all} \& \textbf{you}, be visible when predicting \textbf{need}. However, as the number of layers progresses, tokens like \textbf{Attention} are visible to \textbf{need}, breaking the Markov property.
\begin{figure}[hbt!]
  \centering
  \includegraphics[width=0.5\textwidth]{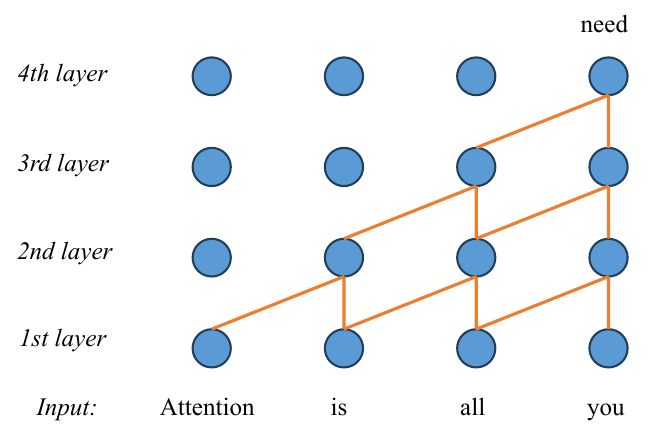} 
  \caption{A second-order attention mask, where the orange lines indicate attention. The input token sequence is [\textbf{Attention}, \textbf{is}, \textbf{all}, \textbf{you}], and the token to be predicted is \textbf{need}.}
  \label{fig: leak}
\end{figure}

\subsection{Training Details}
\label{apd: data}
\paragraph{Loss Function}
The conventional Markov models require global normalization to tackle the label bias problem. 
However, here we cannot perform such normalization because the transition matrix is modeled by a parametric deep neural network which needs traversal of all the possible previous $k$ tokens combination.
After considering the trade-off, we decide to use local normalization as what the vanilla autoregressive transformer does
Thus the loss function is as follows:
\begin{align}
 \mathcal{L} &=- \log P(y_1, y_2, \dots, y_n| X) \nonumber \\
% &=- \log \prod_{i=1}^{n} P(y_i | x, y_{i - k}, \cdots, y_{i - 1}) \notag \\
&=- \sum_{i=1}^{n} \log P(y_i | X, y_{i - k}, \cdots, y_{i - 1}).
\end{align}
Here $k$ is the order of the Markov decoder.

\paragraph{Data Processing}
We learned a BPE model with 32K merge operations for the dataset.
We preprocessed the datasets with a joint BPE~\cite{Sennrich:BPE} with 32K merge operations for En$\Leftrightarrow$De, and 32K bpe for En$\Leftrightarrow$Zh.

\paragraph{Hyperparameters}
For our model and the baselines in our paper, we adopt the Transformer \textsc{Base} architecture, consisting of 6 encoder layers, 6 decoder layers, 8 attention heads, 512 model dimensions, and 2048 hidden dimensions. We use the AdamW optimizer for optimization. To prevent over-fitting, we adopt dropout equals to 0.2. All experiments are conducted on 8 NVIDIA 3090 GPU cards.

\subsection{Previous Markov Models}
\label{apd: related}
\paragraph{Bigram CRF}~\cite{natcrf}. The Bigram CRF employs the Linear-CRF as its decoder while leveraging the standard Transformer Encoder as the encoder part. More specifically, Bigram CRF utilizes a non-autoregressive Transformer decoder to model $P(y_i|x, pos_{i})$. Subsequently, it deploys a low-rank matrix $M \in |V|^2$ to represent the transition probabilities between adjacent tokens, thereby achieving first-order Markov property.
\paragraph{Non-Autoregressive Markov Transformer}~\cite{deng-etal-2021-structure}. This paper utilizes the idea of cascade decoding, beginning with a non-autoregressive model (i.e., zero-order Markov model), and progressively incorporates higher-order Markov dependencies. To accelerate the generation process, it prunes the candidates of the lower-order Markov and also adopts parallel decoding at different positions.
% \subsection{A wrong implementation of Markov Attention Mechanism}
% A natural intuition might suggest that we can directly instill Markovian characteristics within the decoder using a $k$-order attention mask. However, this approach does not truly achieve the desired effect. The underlying complexities of feature dependencies in the multi-layer transformer often surpass the simplistic constraints imposed by such masks. We aim to clarify this perspective in the following discussion.

% The Transformer architecture, particularly when stacked in multiple layers, has been a cornerstone in modern deep learning applications for sequence transduction tasks. One of its key features is the self-attention mechanism, which allows each position in the sequence to focus on different parts of the input based on the learned attention weights.

% When we introduce the $k$-order variant of this self-attention mechanism into the decoding phase, it brings about a convolutional characteristic. Specifically, for a position \( k \) in the target sequence, the context it leverages spans directly over the previous \( 1 + L \times (N-2) \) tokens. Here, \( L \) represents the layer depth in the stacked Transformer decoder.

% This convolutional nature ensures that the context is not limited to just the immediate previous token, as would be the case in a first-order Markov model. Instead, it encompasses a broader range of preceding tokens, making the model's behavior inherently non-Markovian.
\end{document}